# Bottleneck detection by slope difference distribution: a robust approach for separating overlapped cells

ZhenZhou Wang*, Member, IEEE

College of Electrical and Electronic Engineering, Shandong University of Technology, China, 255049.

Corresponding e-mail: zzwangsia@yahoo.com

*Abstract*—To separate the overlapped cells, a bottleneck detection approach is proposed in this paper. The cell image is segmented by slope difference distribution (SDD) threshold selection. For each segmented binary clump, its one-dimensional boundary is computed as the distance distribution between its centroid and each point on the two-dimensional boundary. The bottleneck points of the one-dimensional boundary is detected by SDD and then transformed back into two-dimensional bottleneck points. Two largest concave parts of the binary clump are used to select the valid bottleneck points. Two bottleneck points from different concave parts with the minimum Euclidean distance is connected to separate the binary clump with minimum-cut. The binary clumps are separated iteratively until the number of computed concave parts is smaller than two. We use four types of open-accessible cell datasets to verify the effectiveness of the proposed approach and experimental results showed that the proposed approach is significantly more robust than state of the art methods.

## I. Introduction

BOTTLENECK detection had been proposed for many decades to split the binary clumps (the segmentation blobs) into different parts [1-3]. The principle of bottleneck detection theory is to find the bottleneck points of the binary clump and split the binary clump by connecting two bottleneck points. To this end, many methods have been proposed to detect the correct bottleneck points. In [1], the concavity measure, concavity degree and concavity weight were defined to analyze the concavities and to find the splitting points. In [2], a fast and accurate method was proposed to detect the concavity pixels in the binary clump and the concavity-based rules are defined to generate the candidate split lines. Finally, a merit analysis method was used to select the best split line to separate the binary clump. In [3], a cost function was defined to find the bottleneck points that make the cost function minimum. Although these proposed methods were different, they have one thing in common. All of them require an offline training process to determine a set of parameters, which might be very time-consuming. In recent years, bottleneck detection was combined with ellipse fitting [4] to separate the binary clumps.

This method did not require the training process any longer. It tested much more feature points to see if the separation between any pair of them would yield a best fitted ellipse. The main applications of bottleneck detection is to separate overlapped cells or nanoparticles [1-4]. Due to the great diversities of the overlapping situations of the cells or nanoparticles, the achieved accuracy by the training based bottleneck detection methods [1-3] was below 90%. The bottleneck detection and ellipse fitting combined method [4] achieved 92.2% accuracy, which is still not very high.

An alternative cell or nanoparticle separation method was watershed [5] which has been used in many popular cell segmentation methods or software tools [6]. However, watershed could not overcome the over-segmentation problem which is very severe in most cell or nanoparticle segmentation applications. Similarly, the morphological ultimate erosion [7] would also generate severe over-segmentation problem and thus could not be used directly to separate the overlapped cells. To solve the over-segmentation problem, the authors in [8] proposed the iterative morphological erosion method to separate the overlapping cells or nanoparticles with an area threshold constraint. The iterative morphological erosion method is robust and efficient in separating the overlapping cells that have narrower touching border than the width of the cell [8-9]. However, when the length of the touching line of two overlapping cells or nanoparticles is greater than the widths of the cells or nanoparticles, the iterative erosion method would make one cell disappear before the touching line is eroded off.

Besides the bottleneck detection based methods [1-4], the watershed based methods [5-6] and the morphological erosion based methods [8-9], there are also many other methods proposed in recent years. For instance, the elliptical shape models were used to separate the overlapped cells in [10-12]. Deep learning was used to segment and separate the overlapped cells in [13-16]. The working principle of the elliptical shape modeling methods is to fit the best elliptical shape to the overlapped cells and its implementation codes are available at the website [17] for validation and comparison. On the contrary, the working principles of deep learning vary significantly among different works [13-16]. It is difficult to validate the effectiveness of these deep learning methods since none of the authors disclose their implementation source codes as the authors in [11] did. In addition, the source codes and the implementation demos of the iterative morphological erosion method [8] were available at the website [18] for validation and comparison. Hence, we will compare the approach proposed in this paper with that proposed in [8] and that proposed in [11] extensively.

In this paper, we propose a bottleneck detection approach to separate the cells. Different from the previous research that adopted time-consuming analysis, shape classification and comparison [1-3], we propose to detect the bottleneck points directly by slope difference distribution of the binary clump's boundary. The slope difference distribution has been proposed in [8] to calculate the threshold for robust cell or nanoparticle segmentation and its effectiveness has been validated by recent research. In [19-20], SDD threshold selection has been testified as the most robust method out of sixteen existing methods for segmenting the MRI ventricles. All the source codes and 117 MRI images with ground truths are available at the website [21], which makes it convenient for us compare the proposed method with existing methods. Compared to deep learning method [22] in the quantitative accuracy on the same dataset, SDD threshold selection achieved 92.5% [19] while deep learning method achieved 91% [22]. As pointed out in [19], the SDD threshold selection has the potential to exceed 95% with a different set of parameter and more robust selection method. In [23], SDD clustering was compared with six existing methods and SDD clustering achieved second best with 0.9902 uniformity measure even without parameter calibration. As explained in [8], the optimal performance of SDD is achieved by choosing the optimal calibrated parameters for different types of histograms (or images). In a word, slope difference distribution is effective in computing the threshold points from a histogram. Similarly, the bottleneck points on a boundary are actually the threshold points to separate the boundary. Therefore, the bottleneck detection problem could be simplified by slope difference distribution and no time-consuming analysis, shape classification and comparison are required anymore.

## II. ADVANTAGE AND DISADVANTAGE OF ITERATIVE EROSION

The iterative erosion method proposed in [8] is robust in separating overlapped cells or nanoparticles when the length of the touching line between two cells or nanoparticles is smaller than the width of the cell or nanoparticle. We demonstrate a typical case where iterative erosion works effectively in Fig. 1. Fig. 1 (a) shows the segmentation result of the nanoparticle image by slope difference distribution threshold selection. Fig. 1 (b) shows the output of the iterative erosion with the segmentation result in Fig. 1 (a) as input. The generated seeds by iterative erosion with different labeled color are overlaying on the inputted segmentation result. As can be seen, all the overlapped nanoparticles could be eroded off since their touching line is smaller than the width of the nanoparticle. Experimental results showed that iterative erosion is very effective in separating muscle cells [8] and cells in some SEM images [24].

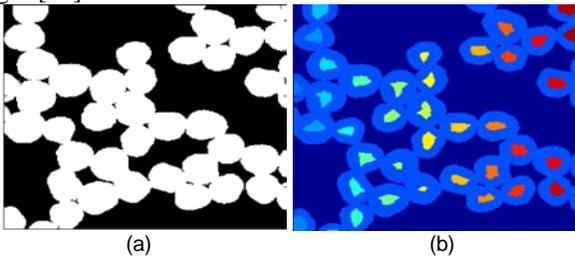

(a)          (b)

**Fig.1. Demonstration of the advantage of iterative erosion**
(**a**) The inputted segmentation result; (**b**) The output of iterative erosion

On the contrary, iterative erosion will not work when the touching line is larger than the width of the cell or nanoparticle. We demonstrate a typical case where iterative erosion does not work in Fig. 2. Fig. 2 (a) shows the segmentation result of the cell image by slope difference distribution threshold selection. Fig. 2 (b) shows the output of the iterative erosion with the segmentation result in Fig. 2 (a) as input. The generated seed in grey is overlaying on the original segmentation result. As can be seen, only one seed is generated for these two overlapped cells and consequently these two overlapped cells could not be separated by iterative erosion.

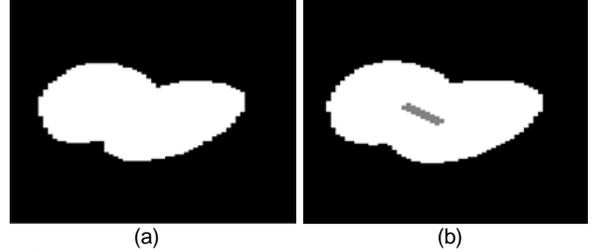

(a)          (b)

**Fig.2. Demonstration of the disadvantage of iterative erosion**
(**a**) The inputted segmentation result; (**b**) The output of iterative erosion

## III. THE PROPOSED APPROACH

### A. Bottleneck points detection by Slope Difference Distribution

The boundary of the binary clump or overlapped cells is in two-dimension while the slope difference distribution calculates the threshold points in one dimension. Therefore, the 2D boundary should be transformed into one dimension at first. The one dimensional boundary is computed as follows.

Firstly, the center of the binary clump is calculated as:

$$(x_c, y_c) = \left( \frac{1}{M} \sum_{i=1}^{M} x_i, \frac{1}{M} \sum_{i=1}^{M} y_i \right) \quad (1)$$

where $(x_i, y_i), i = 1, 2, ..., M$ is the position of $i$th pixel of the binary clump in the image and $M$ is the total number of pixel contained in the binary clump. The points on the exterior boundary of the binary clump are extracted and denoted as $P(x_j, y_j), j = 1, 2, ..., L$. The one dimensional boundary $B_j, j = 1, ..., L$ is then computed as:

$$B_j = \left[ (x_j - x_c)^2 + (y_j - y_c)^2 \right]^{1/2}, j = 1, ..., L \quad (2)$$

Before computing the slope difference distribution of the one dimensional boundary $B_j, j = 1, ..., L$, we filter the boundary in the frequency domain to remove noise at first. The one dimensional boundary, $B_j, j = 1, ..., L$ is transformed into the frequency domain by discrete Fourier transform (DFT).

$$F(k) = \sum_{j=1}^{L} B_j e^{-i\frac{2\pi kj}{L}}; k = 1, ..., L \quad (3)$$

Suppose the low-pass bandwidth of the DFT filter is $W$, the DFT components in Eq. (3) are reduced by the following equation.

$$F'(k) = \begin{cases} F(k); k = 0,1,2,...,W \\ F(k); k = L-W,...,L-1,L \\ 0; k = W+1, W+2,...,L-1-W \end{cases} \quad (4)$$

After component reduction, the boundary distribution is transformed back into spatial domain by the following equation.

$$B_j^s = \frac{1}{L} \sum_{k=1}^{L} F'(k) e^{i\frac{2\pi kj}{L}}; j = 1,...,L \quad (5)$$

where $B_j^s$ is the smoothed boundary. Two lines on both sides of each point on $B_j^s$ are fitted by the following equation.

$$B_j^s = aj + b \quad (6)$$

where $(j, B_j^s); j = N+1, N+2,..., L-N$ is the point on the one dimensional boundary distribution $B_j^s$. $a$ is the slope of the line and $b$ is a constant coefficient. $[a,b]^T$ is computed as:

$$[a,b]^T = (A^T A)^{-1} A^T Y \quad (7)$$

$$A = \begin{bmatrix} j+1-N & 1 \\ j+2-N & 1 \\ \vdots & \vdots \\ j-1 & 1 \\ j & 1 \end{bmatrix} \text{ or } \begin{bmatrix} j & 1 \\ j+1 & 1 \\ \vdots & \vdots \\ j-2+N & 1 \\ j-1+N & 1 \end{bmatrix} \quad (8)$$

$$Y = \begin{bmatrix} B_{j+1-N}^s, B_{j+2-N}^s, ..., B_j^{s'} \end{bmatrix}^T$$
$$\text{or } \begin{bmatrix} B_j^s, B_{j+1}^s, ..., B_{j-1+N}^s \end{bmatrix}^T \quad (9)$$

Two slopes, $a_j^r$ and $a_j^l$ at the $j$th point $(j, B_j^s); j = N+1, N+2,..., L-N$ could be obtained from Eq. 7. The slope difference at the boundary point is computed as:

$$s_j = a_j^r - a_j^l \quad j = 1+N,...,L-N \quad (10)$$

where $s_j$ is the slope difference distribution. Setting the derivative of $s_j$ to zero and solving it, we could get the positions of the valleys $V_i; i = 1,2,...,N_V$ with greatest local variations on the slope difference distribution. The positions $V_i; i = 1,2,...,N_V$ of these valley points are the 1D bottleneck points of the 1D boundary and they also correspond to the indexes of the bottleneck points in the 2D boundary. Accordingly, the set of 2D bottleneck points are computed as:

$$P_i^B = \{(x,y) \mid x = x_{V_i}, y = y_{V_i}, i = 1,2,...,N_V\} \quad (11)$$

We use the binary clump shown in Fig. 2 (a) as an example to demonstrate the process of computing the bottleneck points by slope difference distribution in Fig. 3. Fig. 3 (a) shows the process of detecting the bottleneck points of the 1D boundary by the slope difference distribution. The pink line denotes the 1D boundary computed by Eq. (1)-(5). The green line denotes the slope difference distribution of the 1D boundary. The blue circles denote the detected bottleneck points of the 1D boundary. Fig. 3 (b) shows the bottleneck points (denoted by the blue circles) of the 2D boundary computed by Eq. (11). As can be seen, three bottleneck points are detected while only two of them are true. Therefore, we need additional information to select the correct bottleneck points robustly.

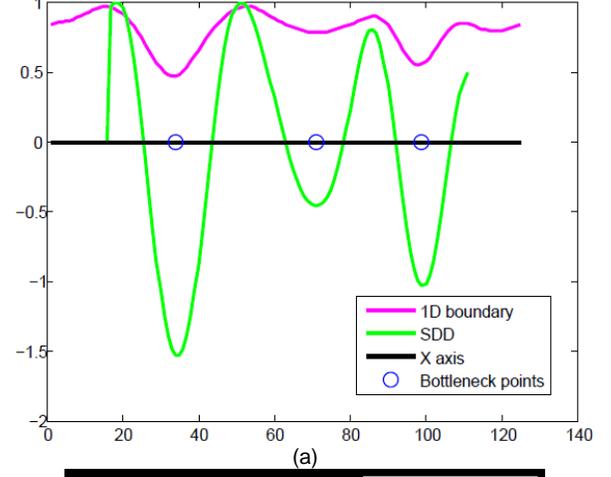

(a)

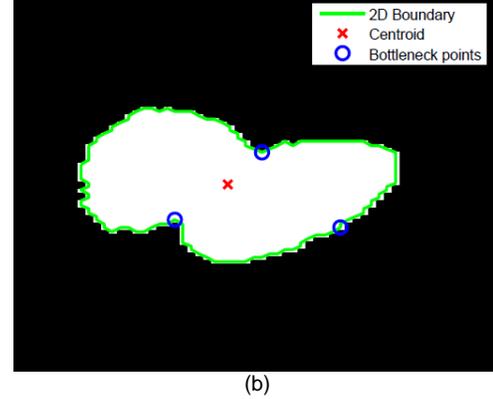

(b)

**Fig.3. Demonstration of the process of computing the bottleneck points by slope difference distribution**
(**a**) The process of detecting the bottleneck points of the 1D boundary by slope difference distribution; (**b**) The extracted 2D boundary and its detected bottleneck points overlaying on the inputted binary clump.

### B. The Proposed Cell Separation Approach

The flowchart of the proposed cell separation approach is shown in Fig. 4 and it contains the following steps.

**Step 1:** The convex hull of the inputted binary clump is calculated as described in [19, 21]. Then, the concave parts of the inputted binary clump are computed by subtracting the inputted binary clump from the generated convex clump. All the concave parts of the inputted binary clump are labeled and the total number of the concave parts is calculated.

**Step 2:** Determine if the binary clump is a single cell or overlapped cell as follows. If the total number of the labeled concave parts $Q$ is less than 2, then the binary clump is determined as a single cell instead of overlapped cells. The binary clump is outputted directly. If the total number of the labeled concave parts $Q$ is greater than 2, the binary clump is determined as overlapped cells. If the binary clump is determined as the overlapped cells, two largest concave parts are selected from all the labeled concave parts.

**Step 3:** If the binary clump is determined as the overlapped cells, then its 2D boundary is extracted and its 1D boundary is computed by Eq. (1)-(5). The 1D bottleneck points are detected by slope difference distribution as described by Eq. (6)-(10). The 2D bottleneck points are computed by Eq. (11).

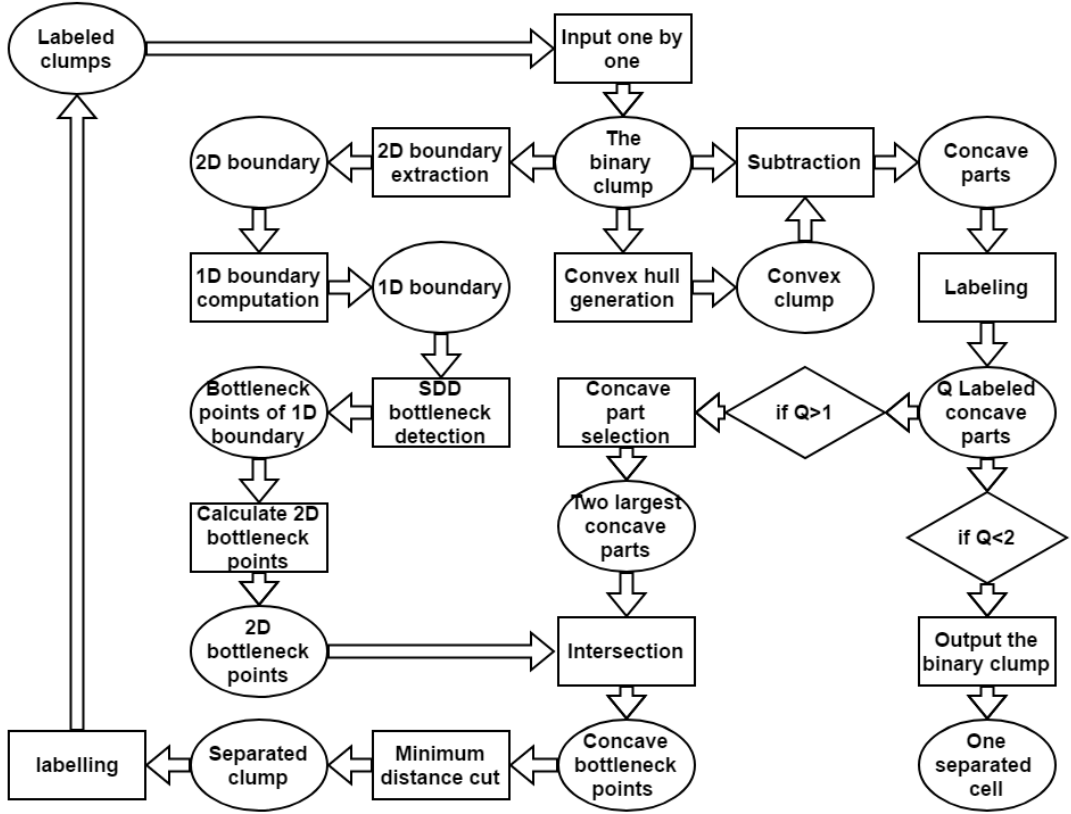

Fig.4. Flowchart of the proposed cell separation approach

**Step 4:** The computed 2D bottleneck points are intersected with the two largest concave parts. If the 2D bottleneck points lie in the concave parts of the inputted binary clump, then they are kept as true bottleneck points. If not, the bottleneck points are deleted.

**Step 5:** The Euclidean distances among the bottleneck points in one concave part and the bottleneck points in the other concave part are calculated. The two bottleneck points that yield the minimum distance are selected as the final two bottleneck points to separate the binary clump. The binary clump is cut into two parts by the touching line between the two final bottleneck points.

**Step 6:** The two separated binary clumps are labeled and inputted as the binary clump for cell separation one by one again. For each separated binary clump, **Steps 1-5** are repeated.

**Step 7: Steps 1-6** are repeated until all the overlapped cells are separated.

For one inputted cell image, the flowchart of identifying all the cells is shown in Fig. 5. Firstly, the cell image is segmented by slope difference distribution threshold selection. All the segmented binary clumps are labeled and then inputted into the proposed cell separation approach one by one. After all the cells are separated, they are labeled again and the coordinate $(x_c^i, y_c^i)$ of the $i$th labeled cell is computed as:

$$(x_c^i, y_c^i) = \left( \frac{1}{N_P} \sum_{j=1}^{N_P} x^i(j), \frac{1}{N_P} \sum_{j=1}^{N_P} y^i(j) \right) \quad (12)$$

where $(x^i(j), y^i(j)), i = 1, 2, ..., N_P$ is the position of the $j$th pixel in the $i$th labeled cell and $N_P$ is the total number of pixel contained in the $i$th labeled cell.

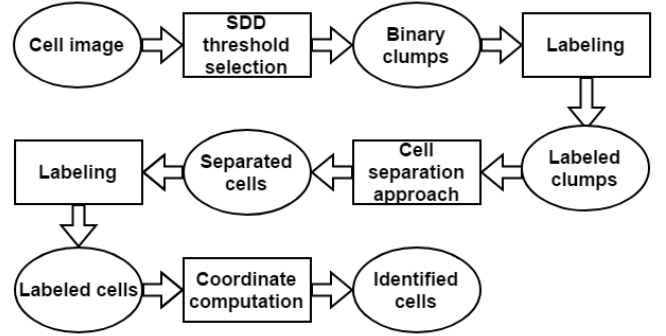

Fig.5. Flowchart of identifying all the cells

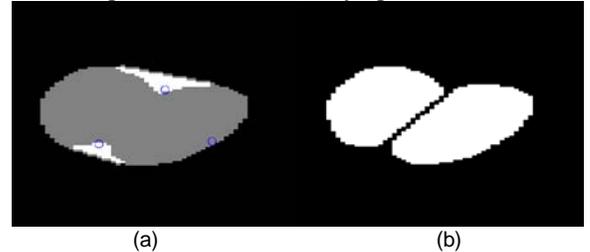

(a)  (b)

**Fig.6. Demonstration of separating the binary clump into two cells by the proposed cell separation approach.** (**a**) The 2D bottleneck points overlaying on the generated convex clump (the bottleneck points lying in the white concave parts are true); (**b**) The separated cells by the split line between the two true 2D bottleneck points.

Fig. 6 shows the cell separation results for the binary clump demonstrated in Fig. 2. In Fig. 6 (a), the white parts are the two largest concave parts of the binary clump and the grey part is the original binary clump. Only the bottleneck points lying in the two largest concave parts are valid. Fig. 6 (b) shows the separated cells by splitting the binary clump with the line that connects the two bottleneck points from two different concave parts. Fig. 7 demonstrates the cell separation process for a complex binary clump. Fig. 7 (a) shows the binary clump segmented by slope difference distribution threshold selection and Fig. 7 (b) shows the detected 2D bottleneck points overlaying on the generated convex clump (Only the bottleneck points overlaying on the two largest concave parts are denoted as true bottleneck points). Fig. 7 (c) shows the corresponding 1D bottleneck points detected by slope difference distribution. Fig. 7 (d) shows the first separation result by splitting the binary clump with the line that connects two bottleneck points from two different concave parts with the minimum Euclidean distance. Fig. 7 (e) shows the detected 2D bottleneck points overlaying on the generated convex clump for the first separated binary clump. Fig. 7 (f) shows the detected 2D bottleneck points overlaying on the generated convex clump for the second separated binary clump. Fig. 7 (g) shows the final separation results for the inputted binary clump.

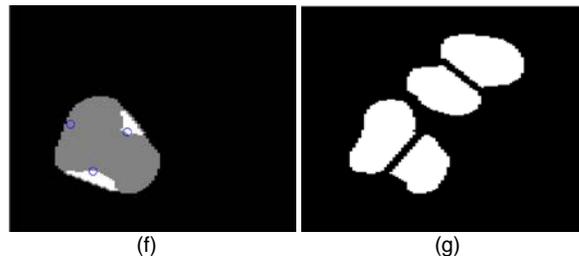

**Fig.7. Demonstration of separating a complex binary clump into four cells by the proposed cell separation approach.** (**a**) The segmented binary clump; (**b**) The 2D bottleneck points overlaying on the generated convex clump (the bottleneck points lying in the white concave parts are true); (**c**) The corresponding 1D bottleneck points detected by slope difference distribution; (**d**) The separated clumps by the split line between the two 2D bottleneck points from different concave parts with minimum distance. (**e**) The 2D bottleneck points of the first separated clump overlaying on the generated convex clump (the bottleneck points lying in the white concave parts are true); (**f**) The 2D bottleneck points of the second separated clump overlaying on the generated convex clump (the bottleneck points lying in the white concave parts are true); (**g**) The final separation results.

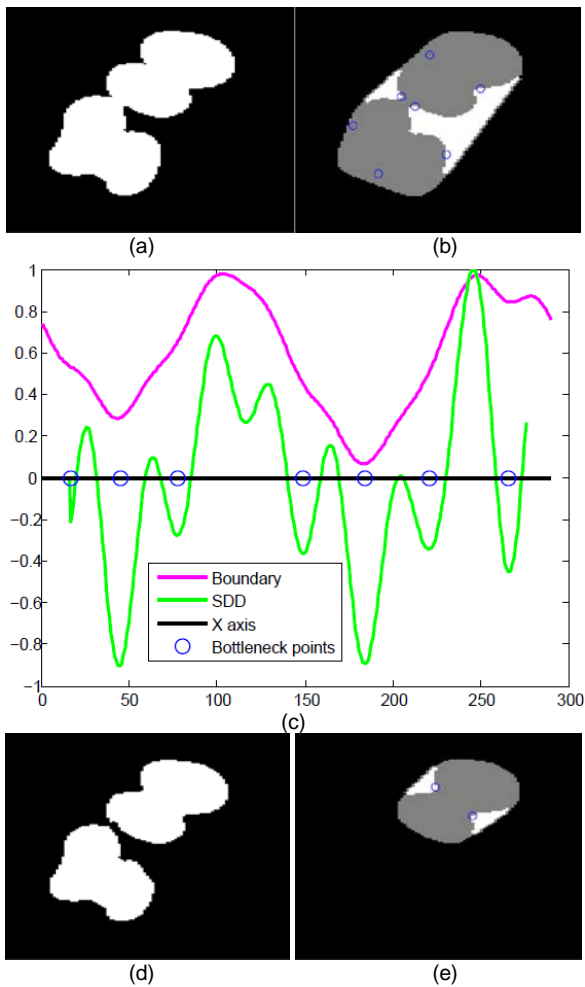

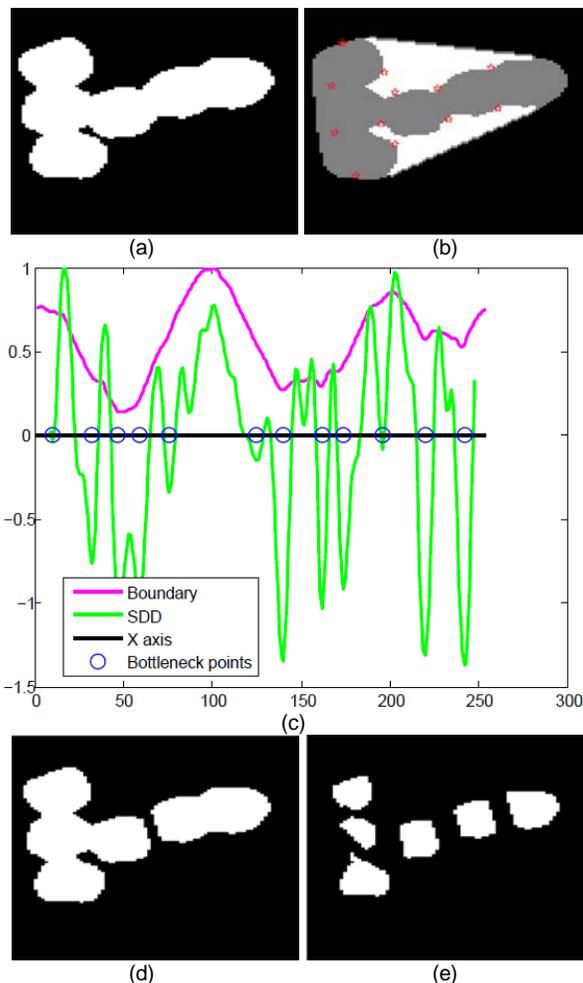

**Fig.8. Demonstration of separating a more complex binary clump into four cells by the proposed cell separation approach.** (**a**) The segmented binary clump; (**b**) The 2D bottleneck points overlaying on the generated convex clump (the bottleneck points lying in the white concave parts are true); (**c**) The corresponding 1D bottleneck points detected by slope difference distribution; (**d**) The first separation result by the split line between the two 2D bottleneck points from different concave parts with minimum distance. (**e**) The final separation results.

Fig. 8 demonstrates the intermediate results of separating a more complex binary clump by the proposed approach. Fig. 8 (a) shows the binary clump segmented by slope difference distribution threshold selection and Fig. 8 (b) shows the detected 2D bottleneck points overlaying on the generated convex clump (Only the bottleneck points overlaying on the two largest concave parts are denoted as true bottleneck points). Fig. 8 (c) shows the corresponding 1D bottleneck points detected by slope difference distribution. Fig. 8 (d) shows the first separation result by splitting the binary clump with the line that connects two bottleneck points from two different concave parts with the minimum Euclidean distance. Fig. 8 (e) shows the final separation results for the inputted binary clump.

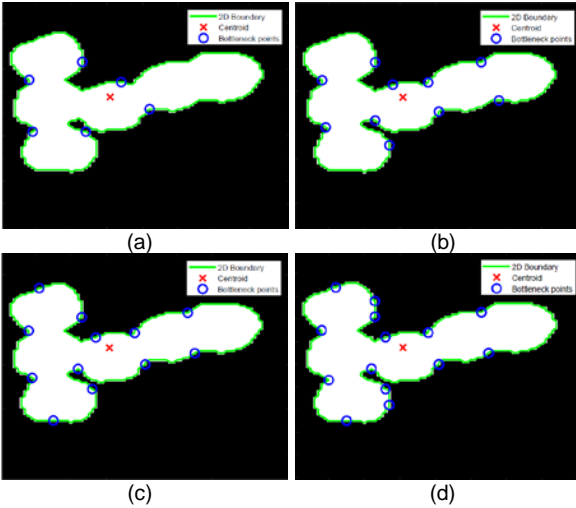

**Fig.9. Demonstration of the effect of selecting different bandwidths for the low-pass DFT filter.** (**a**) The selected bandwidth $W$=10; (**b**) The selected bandwidth $W$=20; (**c**) The selected bandwidth $W$=30; (**d**) The selected bandwidth $W$=80.

Selecting the most effective bandwidth of the low-pass filter formulated in Eq. (4) is critical for the bottleneck detection results. When the bandwidth is too low, most details of the boundary will be removed. When the bandwidth is too large, too many points will be detected. We use the binary clump shown in Fig. 8 to demonstrate the effects of selecting different bandwidths for the low-pass filter. When the bandwidth is selected as $W$=50, the detected bottleneck points are shown in Fig. 8 (b). When the bandwidth is selected as $W$=10, $W$=20, $W$=30 and $W$=80, the detected bottleneck points are shown in Fig. 9 (a), (b), (c) and (d) respectively. As can be seen, some bottleneck points were missing when the bandwidth is selected as 10 and many non-bottleneck points are detected when the bandwidth is selected as 80. The detected bottleneck points with the bandwidths from 20 to 50 are acceptable. To guarantee all the bottleneck points are detected successfully, we select the bandwidth of the low-pass filter as 50 in this paper.

## IV. EXPERIMENTAL RESULTS

### A. Results

We use two open accessible datasets to compare the proposed cell separation approach with state of the art methods. The first dataset is the human osteosarcoma U2OS cell dataset with a total number of forty images (demonstrated in Fig. 6 and Fig. 7). The second dataset is BBBC004_v1_015_images with a total number of twenty synthesized cell images (demonstrated in Fig. 8). Both datasets are available at the website http://www.broadinstitute.org/bbbc [25]. For the state of the art bottleneck detection methods, only the authors in [4] tested their method with the open accessible human osteosarcoma U2OS cell dataset. Therefore, we compared our proposed approach with the method in [4] quantitatively with 40 images of the same type of U2OS cell. Since the source codes in [8] and [11] are conveniently available at the website [17] and [18] respectively, we compared the proposed approach with them both qualitatively and quantitatively on the two datasets.

We use the same evaluation criterion as [4] to compute the quantitative accuracy. The accuracy evaluation measure is called visual accuracy ($VAC$) that is formulated as:

$$VAC = \frac{N_{segment}}{N_{segment} + N_{split} + N_{merge} + N_{add} + N_{missing}} \quad (13)$$

where, $N_{segment}$ denotes the number of cells that are correctly segmented, $N_{split}$ denotes the number of single cells that are incorrectly divided into multiple cells, $N_{merge}$ denotes the number of overlapping cells that are incorrectly identified as single cells, $N_{add}$ denotes the number of cells that are added from background by mistake, $N_{missing}$ denotes the number of cells that are missed. The quantitative comparisons between different methods on the U2OS cell dataset are shown in Table 1. The quantitative comparisons between different methods on the synthetic cell dataset are shown in Table 2. As can be seen, the proposed approach is significantly more accurate than state of the art methods.

Table 1. Comparison of the accuracy of the proposed approach with state of the art methods on the U2OS cell dataset

| Methods | Total Numbers of Cells | $N_{segment}$ | $N_{split}$ | $N_{merge}$ | $N_{add}$ | $N_{missing}$ | VAC |
|---|---|---|---|---|---|---|---|
| [4] | 1452 | 1310 | 18 | 81 | 27 | 16 | 90.22% |
| [11] | 3698 | 3515 | 53 | 86 | 15 | 29 | 95.05% |
| Iterative erosion[8] | 3698 | 3568 | 11 | 97 | 13 | 9 | 96.48% |
| **Proposed approach** | 3698 | **3650** | **9** | **12** | **2** | **6** | **99.22%** |

Table 2. Comparison of the accuracy of the proposed approach with state of the art methods on the synthesized cell dataset

| Methods | Total Numbers of Cells | $N_{segment}$ | $N_{split}$ | $N_{merge}$ | $N_{add}$ | $N_{missing}$ | VAC |
|---|---|---|---|---|---|---|---|
| [11] | 5496 | 5139 | 3 | 328 | 3 | 23 | 93.5% |
| Iterative erosion[8] | 5496 | 5213 | 6 | 233 | 5 | 39 | 94.85% |
| **Proposed approach** | 5496 | **5477** | **3** | **41** | **2** | **3** | **99.11%** |

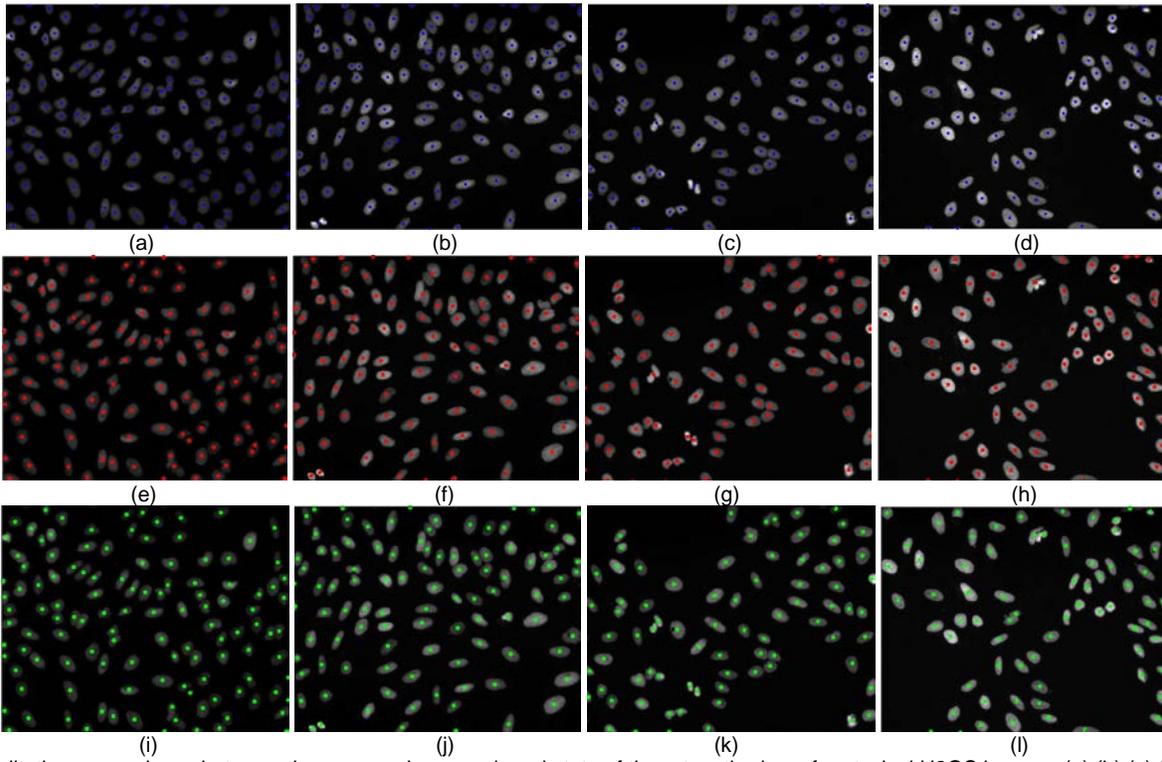

Fig.10. Qualitative comparisons between the proposed approach and state of the art methods on four typical U2OS images. (a),(b),(c),(d), The cell segmentation results by ellipse fitting method [11]. (e),(f),(g),(h), The cell segmentation results by iterative erosion method [8]. (i),(j),(k),(l), The cell segmentation results by the proposed approach.

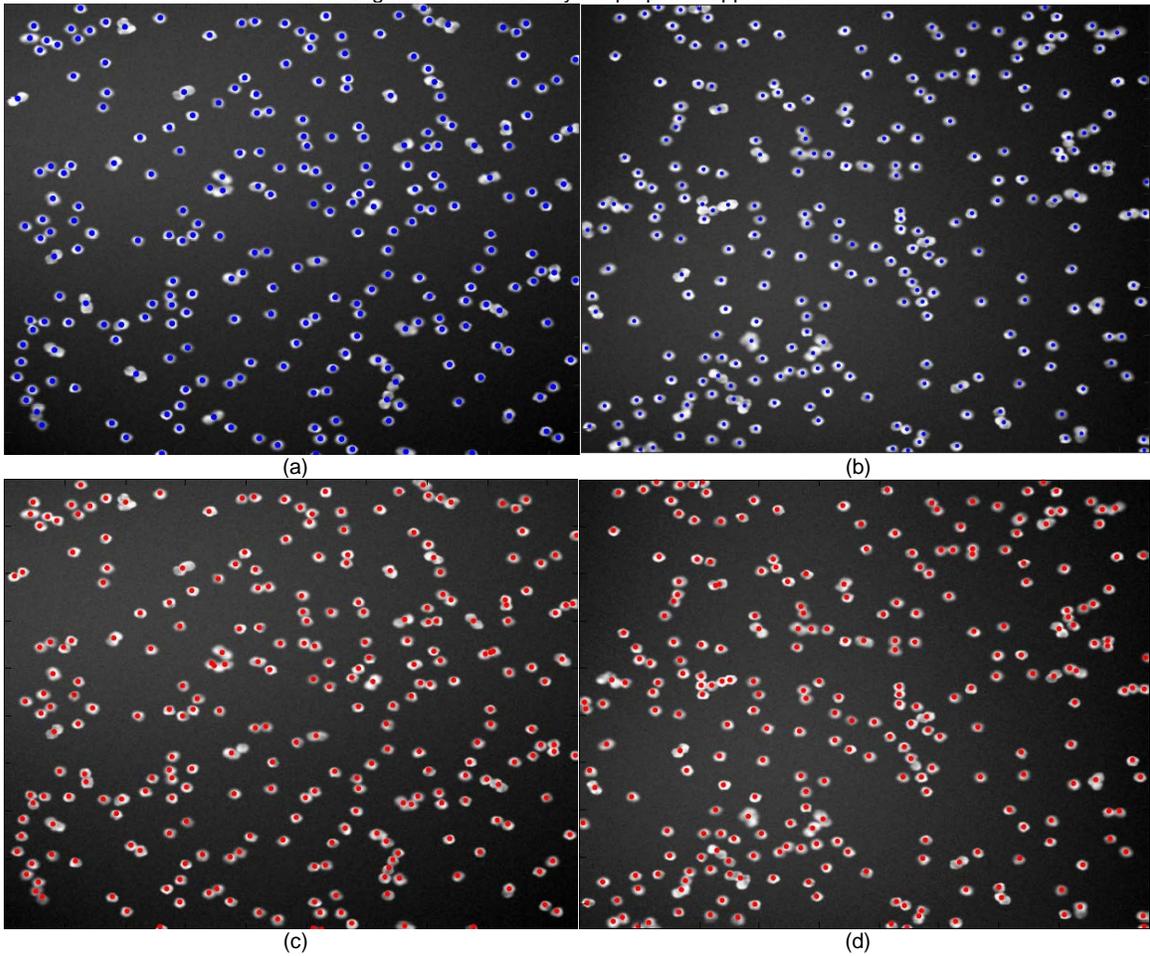

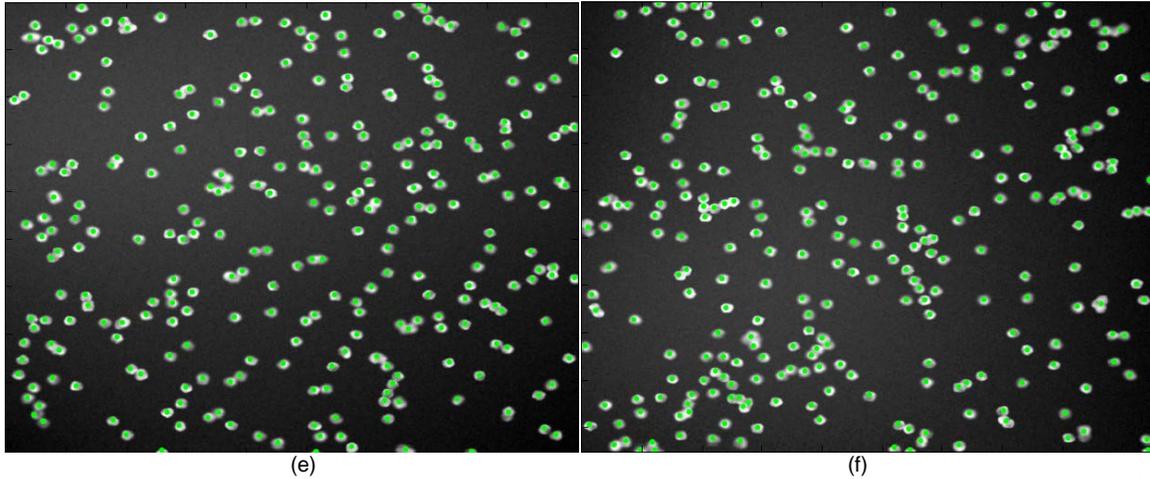

(e)                     (f)

Fig.11. Qualitative comparisons between the proposed approach and state of the art methods on two typical synthetic cell images. (a),(b), The cell segmentation results by ellipse fitting method [11]. (c),(d), The cell segmentation results by iterative erosion method [8]. (e),(f), The cell segmentation results by the proposed approach.

Fig. 10 shows the qualitative comparisons between the proposed approach and state of the art methods [8, 11] in segmenting four typical U2OS cell images. Fig. 11 shows the qualitative comparisons between the proposed approach state of the art methods [8, 11] in segmenting two typical synthetic cell images. As can be seen, the proposed approach is significantly more robust. We show the average computation time used for segmenting the forty U2OS cell images and the twenty synthetic cell images in MATLAB by different methods in Table 3. As can be seen, the computational efficiency of the proposed approach is lower than those of state of the art methods, especially significantly lower than that of the iterative erosion method.

Table 3. Comparison of the average computation time by different methods with the U2OS cell dataset and the synthetic cell dataset.

| Methods | U2OS images | synthetic images |
|---|---|---|
| [4] | 5.8s | **NA** |
| [11] | 36.3 s | 48.3 s |
| Iterative erosion [8] | **0.3 s** | **0.5 s** |
| Proposed approach | 5.6 s | 86.8 s |

Table 4. Comparison of the accuracy of different methods on the Fluo-N2DL-Hela cell dataset [26] and PhC-C2DL-PSC cell dataset [26].

| Methods | Fluo-N2DL-Hela cells | PhC-C2DL-PSC cells |
|---|---|---|
| [11] | 94.4% | 71.2% |
| Iterative erosion [9] | 95.8% | 92.6% |
| **Proposed approach** | **98.9%** | **95.1%** |

In the first dataset, the human osteosarcoma U2OS cells are with organic and irregular shapes although most of them are with ellipse-like shapes. In the second dataset, almost all the synthesized cells are with round or ellipse-like shapes. To show that the proposed method is robust in segmenting cells with organic and irregular shapes, we use another two real cell datasets from [26] to compare the proposed method and state of the art methods in Table 4. As can be seen, the proposed approach was not affected by the organic shapes and its achieved accuracy was significantly better than those of state of the art methods. In addition, the iterative erosion method was not affected by the organic shapes either. On the contrary, the elliptical shape modeling method was affected by the organic shapes greatly. The accuracy of the elliptical shape modeling method for segmenting the PhC-C2DL-PSC cells decreases greatly because the shapes of the segmented cells are not elliptical.

### B. Discussion

Due to the great diversity of cell or nanoparticle overlapping modality and the irregularity of their shapes, it is impossible for a single method to separate all of them both robustly and efficiently. Thus, it is necessary to propose different types of cell separation methods based on the characteristics of the cells. For instance, the iterative erosion method is more effective than the bottleneck detection approaches in segmenting the types of cells or nanoparticles illustrated in Fig. 1 because it takes much less time than the bottleneck detection approaches as demonstrated in Table 3 while its accuracy might be the same as that of the proposed approach. On the other hand, the proposed bottleneck detection approach is significantly more robust than the iterative erosion method in segmenting the U2OS cells, the synthetic cells and many other types of real cells, such as those shown in Table 4.

Similarly, the great diversity of cell or nanoparticle images makes it impossible for a fixed threshold selection scheme to segment all of them robustly. The threshold selection method should be flexible and adjustable. Slope difference distribution (SDD) is first proposed to solve the problem of segmenting different types of images by selecting different threshold points from image histograms. Different from other threshold selection methods, SDD calculates all the candidate threshold points robustly and select the right one to meet the segmentation requirement. In this study, we proved that the bottleneck detection problem could be transformed into a threshold selection problem by transforming the 2D boundary into 1D boundary. As a result, the bottleneck detection problem is simplified. In addition, the robustness of the bottleneck detection is also increased by SDD significantly. For the same open accessible dataset, the ellipse fitting and bottleneck detection combined method [4] only achieved 90.22% accuracy while the SDD based bottleneck detection method achieved 99.22% accuracy. For the bottleneck detection method proposed in [1], only 79.5 % accuracy was achieved for separating 112 clumps while no quantitative accuracy was reported in [2]. For bottleneck detection method proposed in [3], only 87% accuracy was achieved for the blood cells while 92.2% accuracy was achieved for the blood cells by the ellipse fitting and bottleneck detection combined method proposed in [4]. In summary, the advantage of the proposed approach over the previous bottleneck detection methods [1-4] is its significantly better accuracy. The bottleneck points could be detected robustly by SDD and then selected based on the areas of the concave parts to separate the overlapping cells. In addition, the proposed approach does not require any training process while most previous bottleneck detection methods [1-3] need the training process to determine a set of shape parameters.

The elliptical shape modeling methods [10-12] try to find the best fitting ellipses to the overlapped cells and thus will be affected by the shapes of the cells. Their ideal separating conditions are that the cells in elliptical shapes. Based on the qualitative and quantitative comparisons in this paper, the elliptical shape modeling methods [10-12] are more robust than the previous bottleneck detection methods [1-4] in segmenting cells with the elliptical shapes. However, its accuracy deceased significantly when the shapes of the segmented cells are not elliptical as shown in Table 4. For all the four tested datasets in this paper, the accuracy of the elliptical shape modeling method is significantly lower than that of the proposed bottleneck detection approach.

Deep learning has become very popular in cell segmentation in recent years. In [13], deep learning was combined with threshold selection and watershed to segment different types of cells including the U2OS cells, the average accuracy achieved was 86% and the accuracy achieved for the U2OS cell was 85%. In [14], deep learning was used to segment the cervical cells and only 69% accuracy was achieved. In [15], deep learning was combined with threshold selection and morphological operations to segment the nuclei cells and the achieved accuracy was 83%. In [16], deep learning was combined with Cell-profiler to segment the U2OS cells and the achieved accuracy was 81%. As can be seen, the overall accuracy achieved by deep learning based methods was significantly lower than those achieved by other methods [1-12]. In addition, deep learning requires a time-consuming training process. What is worse is that the determined parameters by the training process for one specific type of cells could not be used by other types of cells.

Table 5. Comparison of the typical accuracies and the segmentation conditions of different methods.

| Methods | Typical Accuracy | Conditions |
|---|---|---|
| Previous bottleneck methods [1-3] | <90% for U2OS cells | Training dependent |
| Bottleneck combined with ellipse fitting [4] | 90.22% for U2OS cells | Elliptical shapes Dependent |
| Iterative erosion [8-9] | 96.48% for U2OS cells | None |
| Elliptical modeling [11] | 95.05% for U2OS cells | Elliptical shapes dependent |
| Deep learning [13] | 85% for U2OS cells | Training dependent |
| Proposed bottleneck approach | 99.22% for U2OS cells | None |

To compare different methods more clearly and briefly, we summarize the typical accuracies and the segmentation conditions of different methods in table 5. In summary, the major contributions of this paper include: (1), a slope difference distribution based method is proposed to detect the bottleneck points and its accuracy is significantly more accurate than those of the previous bottleneck detection methods [1-4]. (2), the effectiveness of the slope difference distribution in detecting the threshold points is verified further in 2D boundaries. (3), a bottleneck detection approach is proposed to segment different types of cells and its accuracy is significantly better than those of state of the art methods based on the quantitative comparisons on four different types of cell datasets.

## V. CONCLUSION

In this paper, a bottleneck detection approach is proposed to separate the overlapped cells robustly. The bottleneck points are detected by slope difference distribution directly and thus the off-line training process is not required any more. On the contrary, the previous bottleneck detection methods [1-3] rely on the training process greatly. Compared to the bottleneck detection and ellipse fitting combined method [4] or the elliptical modeling methods [10-12], the proposed approach is suitable for segmenting cells with any organic shapes while the elliptical methods are only suitable for segmenting cells with elliptical shapes. Compared to the iterative erosion method [8-9], the proposed approach is not limited by the length of the touching line between two overlapping cells. The proposed approach is compared with state of the art methods both qualitatively and quantitatively on four types of open accessible cell datasets. Experimental results show that the proposed approach is significantly more robust than state of the art methods.